\documentclass[11pt,a4paper]{article}
\usepackage[hyperref]{AACL-IJCNLP2020}
\usepackage{times}
\usepackage{latexsym}

\usepackage{microtype}

\usepackage{caption}
\usepackage{url}
\usepackage{algorithm}
\usepackage{algorithmic}
\usepackage{graphicx}
\usepackage{amsmath}
\usepackage{amsfonts}
\usepackage{multirow}

\usepackage{amsmath,amsfonts,amsthm}
\usepackage{bbm,bm}
\usepackage{graphicx}
\usepackage{color}

\aclfinalcopy 

\title{{N}eural {L}atent {D}ependency {M}odel for {S}equence {L}abeling}

\author{YangZhou$^{\diamond *}$, Yong Jiang$^{\dagger}$\thanks{\hspace{1mm} Yang Zhou and Yong Jiang are the first author.}  , Zechuan Hu$^{\diamond}$,  Kewei Tu$^{\diamond }$  \\
 $^\diamond$School of Information Science and Technology, ShanghaiTech University \\
 Shanghai Engineering Research Center of Intelligent Vision and Imaging \\
 Shanghai Institute of Microsystem and Information Technology, Chinese Academy of Sciences \\
 University of Chinese Academy of Sciences \\
 $^\dagger$DAMO Academy, Alibaba Group \\
  {\tt \{zhouyang1,huzch,tukw\}@shanghaitech.edu.cn} \\
  {\tt \{yongjiang.jy\}@alibaba-inc.com} \\
 }

\date{}

\begin{document}
\maketitle
\begin{abstract}
Sequence labeling is a fundamental problem in machine learning, natural language processing and many other fields. A classic approach to sequence labeling is linear chain conditional random fields (CRFs). When combined with neural network encoders, they achieve very good performance in many sequence labeling tasks. One limitation of linear chain CRFs is their inability to model long-range dependencies between labels. High order CRFs extend linear chain CRFs by modeling dependencies no longer than their order, but the computational complexity grows exponentially in the order. In this paper, we propose the Neural Latent Dependency Model (NLDM) that models dependencies of arbitrary length between labels with a latent tree structure.
We develop an end-to-end training algorithm and a polynomial-time inference algorithm of our model. We evaluate our model on both synthetic and real datasets and show that our model outperforms strong baselines.
\end{abstract}

\section{Introduction}
Sequence labeling is the problem of labeling each token of an input sequence to produce an output sequence of labels. It is a fundamental problem in many fields such as machine learning, natural language processing, computer vision, and speech recognition. 

To solve sequence labeling, the linear chain conditional random field (CRF) \cite{lafferty2001conditional} is developed which assumes dependency relations between neighboring labels conditioned on the input sequence. Using hand-crafted feature representations of the input sequence, linear chain CRFs have been an important benchmark for many real world sequence labeling tasks \cite{lafferty2001conditional,sha2003shallow,gimpel2011part}.  The recent renaissance in neural networks has led to significant progress on automatic feature extraction with deep neural networks. By modeling the potential functions in CRFs with deep neural networks, neural CRF models achieve significant performance boost over traditional feature-based CRFs on many tasks \cite{collobert2011natural,chen2015long,ma2016end,lample2016neural}.

In many applications, however, there exist long-range dependency relations between output labels. To model such long-range dependencies, the potential function of a CRF can be factorized by cliques containing more than two neighboring labels, resulting in a high order CRF. A major disadvantage of high order CRFs is that the computational complexity of training and decoding is exponential in the order. 

In this paper, we present a novel latent variable model called the Neural Latent Dependency Model (NLDM) that solves the long-range dependency problem by modeling the dependency relations between output labels with a latent tree structure. Similar to neural CRFs, our model employs a deep neural network (a Bi-LSTM) for automatic feature extraction from the input sequence.
Unlike high order CRFs, our model does not place an upper bound to the dependency length between labels. By not allowing edge crossing in the latent tree structure, our model can be trained in an end-to-end way in polynomial time with respect to the sequence length and the label number. Although exact decoding (predicting the output labels given an input sequence) may not be tractable in our model, we propose an efficient approximate decoding algorithm.

To empirically verify the benefits of capturing long-range dependencies, we generated synthetic datasets from a neural \emph{infinite order} generative model and show that our model consistently outperforms the neural CRF on these datasets. We then empirically show that our model also outperforms the neural CRF and other baselines on nine real datasets of sequence labeling. 

\section{Background}
\label{sec:bg}
\subsection{Sequence Labeling with Linear Chain CRFs} \label{sec:2.1}
Given an input sequence $\mathbf{x}=x_1, x_2, ..., x_n$ of length $n$ and an output label sequence $\mathbf{y}=y_1, y_2, ..., y_n$ where $y_i$ is the label of input token $x_i$, a conditional random field model \cite{lafferty2001conditional} is used to assign a score $\phi(\mathbf{x}, \mathbf{y})$ to the pair $(\mathbf{x}, \mathbf{y})$. Linear chain CRFs make the so-called first-order assumption that given the input sequence $\mathbf{x}$ and a label $y_i$, the labels to the left of $y_i$ are independent of the labels to the right of $y_i$. Therefore, the score $\phi (\mathbf{x}, \mathbf{y})$ can be factorized as follows,

\[\phi(\mathbf{x}, \mathbf{y})=\sum_{i=1}^{n} \left(\phi (\mathbf{x}, y_i, i) + \psi(y_{i-1}, y_{i})  \right)\]
In the equation, $\phi(\mathbf{x}, y_i, i)$ is the emission score, which is the score of a particular label $y_i$ given sequence $\mathbf{x}$ at position $i$. 
$\psi(y_{i-1}, y_{i})$ is the transition score, which is the score of the transition from label $y_{i-1}$ to label $y_i$. With $m$ possible labels for each token, the transition score is parameterized with matrix $\mathbf{t} \in \mathbb{R}^{m\times m}$.


To train the model, we typically aim to minimize the regularized negative conditional log-likelihood of the training data, which is,
\begin{align*}
& J(\boldsymbol{\Theta}) = -\sum_{\alpha=1}^{N} \log P(\mathbf{y}^{\alpha} | \mathbf{x}^{\alpha}) + \Omega ||\mathbf{\Theta}||_2^2 \\
&= -\sum_{\alpha=1}^{N} \left( \Phi(\mathbf{x}^{\alpha}, \mathbf{y}^{\alpha})-\log \mathbf{Z} (\mathbf{x}^{\alpha}) \right) + \Omega ||\mathbf{\Theta}||_2^2 
\end{align*}
where $\Omega$ is a hyper-parameter, and $\mathbf{Z}(\mathbf{x}^{\alpha})$ is the partition function, which can be efficiently calculated using the forward algorithm with time complexity $O(n \times m^2)$.


\subsection{Sequence Labeling with Deep Neural Networks} \label{sec:2.2}
Linear chain CRFs have been extremely popular in sequence labeling for their simplicity and efficiency. 
However, hand-crafted feature engineering in CRFs requires domain-specific knowledge and task-specific resources. 

Recent progress in deep learning enables effective automatic feature extraction for many sequence labeling tasks in natural language processing, including word segmentation \cite{chen2015long}, named entity recognition \cite{ma2016end,lample2016neural}, POS tagging \cite{ma2016end} and so on. All these approaches employ a non-linear deep neural network that takes as input pre-trained word embeddings \cite{mikolov2013efficient,pennington2014glove} and outputs a continuous representation for each word in the input sequence that captures not only the word information but also information of its context.

Since the neural network already captures sufficient context information of the input tokens, it can be used with or without the original CRF model for sequence labeling. Mathematically, after the hidden representation $\mathbf{h}_i \in \mathbb{R}^{D_{h}}$ is produced by the deep neural network for each token $x_i$ at position $i$, the emission score can be computed using a bilinear function,
\[\phi(\mathbf{x}, y_i, i) = \mathbf{h}_{i}^{T} \mathbf{W}_{e} \mathbf{t}_{y_{i}}\]
where $\mathbf{t}_{y_{i}} \in \mathbb{R}^{D_l}$ is the embedding for label $y_i$, $\mathbf{W}_{e} \in \mathbb{R}^{D_h\times D_l}$ is a weight matrix.
The label for each token can then be predicted independently by maximizing the following conditional probability.
\[P(\mathbf{y}|\mathbf{x}) = \prod_{i=1}^{n} P(y_i|\mathbf{x})=\prod_{i=1}^{n} \frac{e^{\phi(\mathbf{x}, y_i, i)}}{\sum_{y'} e^{\phi(\mathbf{x}, y', i)}}\]

We call this approach neural softmax. On the other hand, we can also explicitly model label correlation by using a CRF on top of the output representation of the neural network, resulting in the so-called neural CRF model with the following conditional probability:
\[P(\mathbf{y}|\mathbf{x})=\frac{e^{\sum_{i=1}^{n} \big( \phi(\mathbf{x}, y_i, i) + \psi (y_{i-1}, y_{i})\big)}}{\sum_{\mathbf{y'}}e^{\sum_{i=1}^{n} \big( \phi(\mathbf{x}, y'_i, i) + \psi (y'_{i-1}, y'_i)\big)}}\]

In the literature, for the neural softmax model, \newcite{collobert2011natural} proposed a simple yet effective feed-forward neural network for predicting the label for each token by encoding its context information within a fixed window size. For the neural CRF model, \newcite{huang2015bidirectional} was the first to show the promise of applying a (bi-directional) Long-Short-Term-Memory network (LSTM) together with a CRF layer in sequence labeling tasks. \newcite{lample2016neural} enhanced word embeddings with character-based word representations produced by a Bi-LSTM network and showed its success in NER tasks for many languages. \newcite{ma2016end} employed a convolutional neural network (CNN) to represent the morphological information of a word from character embeddings in POS tagging and NER tasks.

\subsection{High Order CRFs}
Although linear chain CRFs have been shown to be effective in sequence labeling, they only model dependencies between neighboring labels and therefore have limited capacity in capturing long-range dependencies. High order CRFs extend linear chain CRFs by explicitly modeling all the label dependencies no longer than their order. However, the computational complexity of training and inference in high order CRFs grows exponentially in the order \cite{ye2009conditional}. For example, the time complexity of exact inference in a $k$-th order CRF is $O(n\times m^{k+1})$. When $k$ is large, the computational cost is not affordable.

Several approaches have been proposed to alleviate this problem under different conditions. \newcite{ye2009conditional} showed that by restricting the number of distinct label sequences used in the features, the time complexity for decoding can be reduced to polynomial. \newcite{qian2009sparse} proposed a sparse higher order model in which learning and inference with high order features can be efficient under the pattern sparsity assumption. 

Compared with these approaches, 
our model achieves efficient learning and inference without any assumption on the features or the properties of the output labels. 

\subsection{Dependency Grammars and Its Variants}
\label{sec:dg}
Our approach is closely related to dependency grammars.
A dependency grammar is a simple yet effective way of covering syntactic relations between words in a sentence in the natural language processing field \cite{nivre2005dependency}. The syntactic relations form a dependency parse tree of the sentence. Given a sentence, discovering its parse tree is called dependency parsing. There are two types of dependency grammars that differ in whether the dependencies in the parse tree are allowed to cross or not. Crossing is allowed by non-projective dependency grammars and is not allowed by projective dependency grammars.
Note that a dummy root node is typically added at the beginning of the sentence which connects to the actual root word of the parse tree. If multiple edges are allowed to emit from the dummy root node, then we have a dependency forest over the sentence.

The quality of a dependency parse tree $\mathbf{z}$ (represented by a set of directed dependency edges) over a sequence of tokens $\mathbf{x}$ is measured by a score function $S(\mathbf{x}, \mathbf{z})$. One simple way of formulating the score function is to decompose the score by dependency edges in the tree, which is called first-order dependency factorization:
\[S(\mathbf{x}, \mathbf{z})=\sum_{(i,j) \in \mathbf{z}} s(\mathbf{x}, i, j)\]
where $s(\mathbf{x}, i, j)$ is defined as the score of the edge $(i,j)$.
For first order projective dependency parsing, \newcite{eisner-1996-coling} proposed an efficient dynamic programming algorithm for supervised parsing and \newcite{paskin2001cubic} proposed an inside-outside algorithm for unsupervised parsing. \newcite{mcdonald2005non} proposed a first order non-projective dependency parsing algorithm based on maximum spanning trees (MST) in directed graphs.


Our model draws inspiration from these approaches. However, instead of modeling the dependency relations between tokens of the input sequence, we model the relations between labels in the output label sequence. 

\subsection{Latent Tree Models for Downstream Tasks} 
Besides sequence labeling, latent tree models are useful in many other applications. \newcite{yogatama2016learning} proposed to learn task-specific constituency tree structures for natural language sentences with the REINFORCE algorithm \cite{williams1992simple}. Similarly, \newcite{choi2017unsupervised} proposed to learn task-specific constituency trees by employing the Gumbel-softmax trick \cite{jang2016categorical}. \newcite{williams2017learning} compared the two approaches empirically and found that these training methods lead to inconsistent trees of natural language sentences and the Gumbel-softmax trick leads to better classification accuracy than the REINFORCE algorithm.  There also exist previous work on latent dependency tree models that utilizes structured attention mechanisms \cite{kim2017structured,liu2017learning} for sentence and document classification problems.

\section{Neural Latent Dependency Model}
\begin{figure}[t]
	\centering
	\includegraphics[scale=0.2]{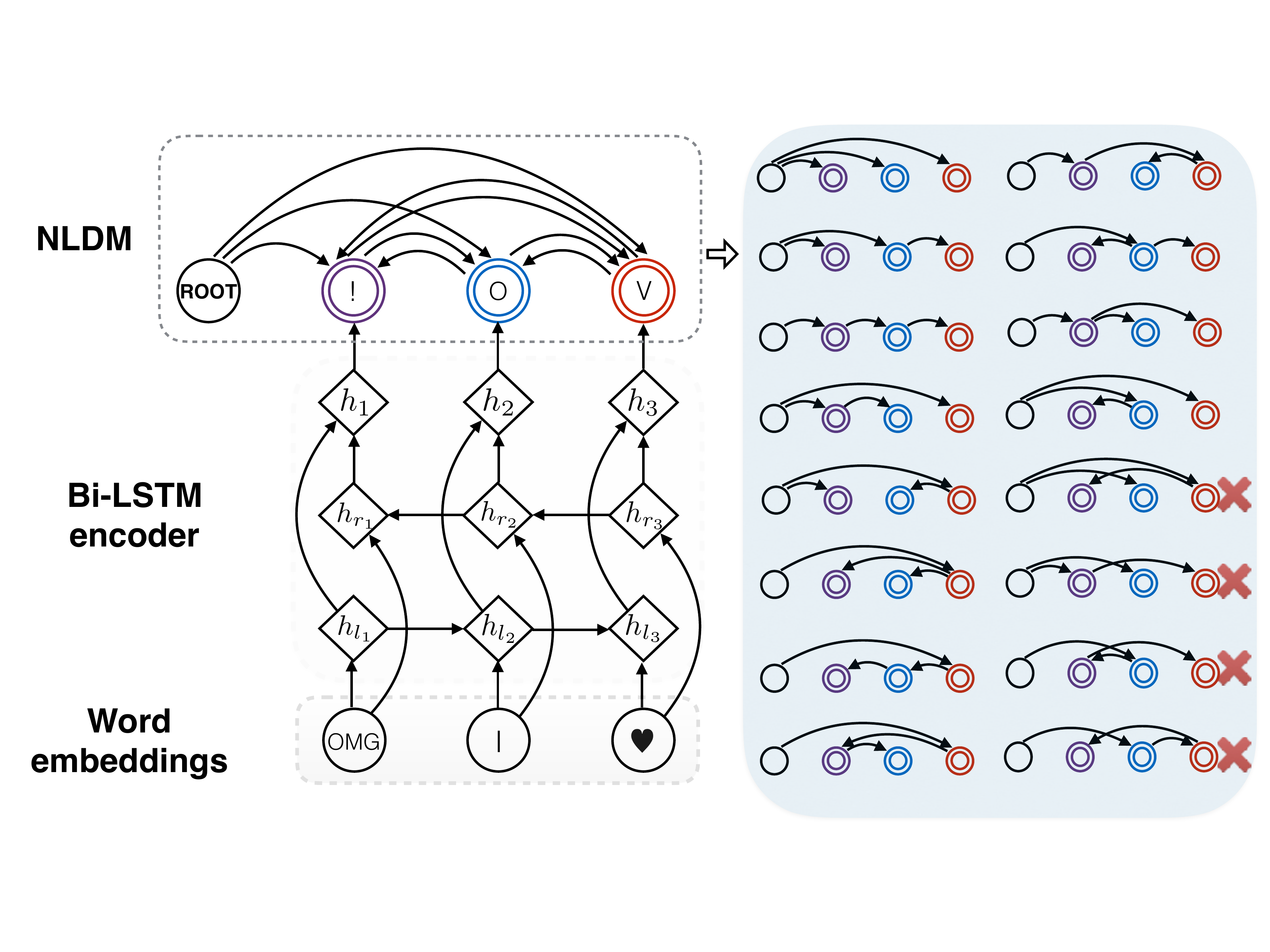}
	\caption{\textbf{Left}: our model architecture for the input sequence \text{``OMG I love"} with the output label sequence \text{``!, O, V"}. \textbf{Right}: the set of 16 possible dependency trees, among which 12 are projective trees that are considered by our model.}
	\label{fig:network}
\end{figure}

The neural latent dependency model can be seen as an extension of linear chain neural CRFs in which the dependency relations between the labels form a projective dependency tree instead of a linear chain. 
The dependency tree has a dummy root node and we allow multiple edges connected to the dummy root. Given a label sequence with multiple tokens, there is more than one possible dependency tree and we regard the tree structure as a latent variable that shall be marginalized during inference. Figure \ref{fig:network} shows our model architecture with an example input sequence.

\subsection{Model} \label{sec:model}
\paragraph{Input Representation} Each token in the input sequence is represented by a continuous embedding $\mathbf{x}_i \in \mathbb{R}^{D_x}$. A bi-directional recurrent neural network (RNN) \cite{rumelhart1986learning,hochreiter1997long} is applied on top of the embeddings to produce a vector for each token that captures context information of the token in the sequence. The bi-directional RNN produces two hidden vectors $\mathbf{h}_{lt}$ and $\mathbf{h}_{rt}$ for the left-to-right RNN and the right-to-left RNN seperately.

We concatenate the two hidden vectors from the two RNNs as our final representation for the $t$-th token $\mathbf{h}_t = [\mathbf{h}_{lt}; \mathbf{h}_{rt}] \in \mathbb{R}^{2D_{h}}$. 

\paragraph{Edge Scoring} Given two positions $i$ and $j$ in a sequence $\mathbf{x}$ and its label sequence $\mathbf{y}$, we can calculate the score of a dependency edge from label $y_i$ to label $y_j$ as,
\begin{align*}
s(\mathbf{x}, \mathbf{y}, i, j) &= \phi (\mathbf{x}, y_j, j) + \psi(y_{i}, y_{j}) \\
&= \mathbf{h}_{j}^{T} \mathbf{W}_{e} \mathbf{t}_{y_{j}} + \bm{\psi}^{dir}_{y_i y_j}
\end{align*}
where $\phi (\mathbf{x}, y_j, j)$ is the emission score at position $j$, $\psi (y_{i}, y_{j})$ is the transition score of the dependency edge, $\mathbf{t}_{y_{j}}$ is the embedding of label $y_{j}$, $\mathbf{W}_{e}$ is a weight matrix, and $\bm{\psi}^{dir}$ is a direction-specific transition matrix chosen based on the direction of the dependency edge. Note that the dummy root node is at position 0 and has a fixed dummy label.

Instead of using a transition matrix $\bm{\psi}^{dir}$, we may also model the transition score with a bilinear function over label embeddings. Further, we may merge the emission and transition scores and compute the edge score with a trilinear function over the embeddings of the token and two labels, which is more expressive than the original score function,
\begin{align*}
    s(\mathbf{x}, \mathbf{y}, i, j) &= \mathbf{h}_{j}^{T} \mathbf{U} \mathbf{t}_{y_{i}} \mathbf{t}_{y_{j}}
\end{align*}
where $\mathbf{U}$ is a order-3 tensor. To reduce computational complexity, we assume that $\mathbf{U}$ has rank of $D_r$ and can be represented as the product of three matrices $\mathbf{U}_1\in \mathbb{R}^{D_r \times D_h}$, $\mathbf{U}_2$ and $\mathbf{U}_3$ $\in \mathbb{R}^{D_r \times D_l}$, 
\begin{align*}
    & s(\mathbf{x}, \mathbf{y}, i, j) = \sum_{k=1}^{D_r} (\mathbf{v}_1 \circ \mathbf{v}_2 \circ \mathbf{v}_3)_k \\
    & \mathbf{v}_1 = \mathbf{U}_{1}\mathbf{h}_i;\quad\mathbf{v}_2 =  \mathbf{U}_{2}\mathbf{t}_{y_{i}};\quad\mathbf{v}_3 =  \mathbf{U}_{3}\mathbf{t}_{y_{j}};\quad
\end{align*}
where $\circ$ is the element-wise product.

\paragraph{Tree Scoring} Given an input sequence $\mathbf{x}$ and an output label sequence $\mathbf{y}$, the score of a dependency tree $\mathbf{z}$ over the labels (represented by a set of directed dependency edges) is factorized by edges:
\[s(\mathbf{x}, \mathbf{y}, \mathbf{z}) = \sum_{(i, j) \in \mathbf{z}} s(\mathbf{x}, \mathbf{y}, i, j)\]
This factorization is motivated by first order dependency parsing in natural language processing \cite{eisner-1996-coling,paskin2001cubic,mcdonald2005non}. We require that the dependency tree is projective, which means the edges in the tree cannot cross each other.

\paragraph{Label Sequence Scoring} In most applications, the dependency relations between output labels are not visible and therefore the dependency tree $\mathbf{z}$ can be seen as a latent variable. For a given input sequence $\mathbf{x}$ and an output label sequence $\mathbf{y}$, their score can be computed by marginalizing the latent variable $\mathbf{z}$:
\[S(\mathbf{x}, \mathbf{y}) = \sum_{\mathbf{z}} e^{s(\mathbf{x}, \mathbf{y}, \mathbf{z})} = \sum_{\mathbf{z}} e^{\sum_{(i,j) \in \mathbf{z}} s(\mathbf{x}, \mathbf{y}, i, j)}\]
or can be approximated by the maximum score over the latent variable $\mathbf{z}$.
\[S(\mathbf{x}, \mathbf{y}) \approx \max_{\mathbf{z}} e^{s(\mathbf{x}, \mathbf{y}, \mathbf{z})} = \max_{\mathbf{z}} e^{\sum_{(i,j) \in \mathbf{z}} s(\mathbf{x}, \mathbf{y}, i, j)}\]
The summation in the first equation can be efficiently computed using the inside algorithm \cite{baker1979trainable} and the maximization in the second equation can be efficiently computed using the CYK algorithm \cite{eisner-1996-coling}. Both algorithms have $O(n^3)$ time complexity and their running time can be significantly improved with vectorized operations in GPUs. Note that the original versions of the two algorithms output a dependency tree over the sequence of tokens, but here we run them to produce a tree over the sequence of labels.

\paragraph{Partition Function} We can also perform marginalization over $\mathbf{y}$ to compute the partition function,
 \[\mathbf{Z}(\mathbf{x}) = \sum_{\mathbf{y}} S(\mathbf{x}, \mathbf{y})=\sum_{\mathbf{y}} \sum_{\mathbf{z}} e^{s(\mathbf{x}, \mathbf{y}, \mathbf{z})}\]
We can efficiently compute the partition function using a modified inside algorithm. The basic idea behind the algorithm is that we regard our model as a weighted context-free grammar (WCFG) that contains two layers: the first layer is a projective dependency grammar (a subclass of WCFG) over the label sequence and the second layer is a lexicon that contains unary production rules in the form of a label producing a token. The inside algorithm on this WCFG computes the partition function of our model. The time complexity is $O(m^2 \times n^3)$ ($m$ is the number of possible labels for each token) and the running time can be improved with vectorized operations in GPUs. 

There are two main reasons as to why we choose to model the dependencies between the output labels with a projective tree rather than a non-projective tree. One reason is that a projective tree contains more short dependencies which are desired in many sequence labeling problems. The other reason is that computing the partition function for non-projective trees is $\#$P-hard.

\subsection{Inference}
At test time, given an input token sequence $\mathbf{x}$, we aim to discover the output label sequence with the highest score,
\[\hat{\mathbf{y}}=\arg\max_{\mathbf{y}} S(\mathbf{x}, \mathbf{y})=\arg\max_{\mathbf{y}} \sum_{\mathbf{z}} e^{s(\mathbf{x}, \mathbf{y}, \mathbf{z})}\]
This problem is known as the Maximum A Posteriori problem \cite{murphy2012machine}, which is very difficult in general. Instead of trying to find an exact solution, we propose to solve this problem approximately,
\begin{align*}
\hat{\mathbf{y}}=\arg\max_{\mathbf{y}} S(\mathbf{x}, \mathbf{y})  &\approx \arg\max_{\mathbf{y}} \max_{\mathbf{z}} e^{s(\mathbf{x}, \mathbf{y}, \mathbf{z})} \\
&=\arg\max_{\mathbf{y}} \max_{\mathbf{z}} s(\mathbf{x}, \mathbf{y}, \mathbf{z})
\end{align*}
The last argmax in the equation can be exactly solved in polynomial time by simply replacing the \textbf{sum} operation with the \textbf{max} operation in our modified inside algorithm. 

\subsection{Learning}
Given a training set of sequences $\mathbf{x}^1, \mathbf{x}^2, ..., \mathbf{x}^N$ and their label sequences $\mathbf{y}^1, \mathbf{y}^2, ..., \mathbf{y}^N$, the learning objective function is the regularized conditional log likelihood:
\begin{align*}
& J(\mathbf{\Theta}) = -\frac{1}{N}\sum_{\alpha = 1}^{N} \log P(\mathbf{y}^{\alpha} | \mathbf{x}^{\alpha}) + \Omega ||\mathbf{\Theta}||_2^2 \\
& =-\frac{1}{N}\sum_{\alpha = 1}^{N} \Big( \log S(\mathbf{x}^{\alpha}, \mathbf{y}^{\alpha}) - \log \mathbf{Z}(\mathbf{x}^{\alpha}) \Big)  + \Omega ||\mathbf{\Theta}||_2^2
\end{align*}
where $\mathbf{\Theta}$ contains the parameters of the LSTM and the edge score function, and $\Omega$ is a hyperparameter.
The gradient of $\log S(\mathbf{x}^{\alpha}, \mathbf{y}^{\alpha})$ can be computed via the higher-dimensional chain rule,
\begin{align*}
& \frac{\partial \log S(\mathbf{x}^{\alpha}, \mathbf{y}^{\alpha})}{\partial \mathbf{\Theta}} \\
& = \sum_{i} \sum_{j} \underbrace{\frac{\partial \log S(\mathbf{x}^{\alpha}, \mathbf{y}^{\alpha})}{\partial s(\mathbf{x}^{\alpha}, \mathbf{y}^{\alpha}, i, j)}}_{\text{term 1}} \underbrace{\frac{\partial s(\mathbf{x}^{\alpha}, \mathbf{y}^{\alpha}, i, j)}{\partial \mathbf{\Theta}}}_{\text{term 2}}
\end{align*}

It can be shown that term 1 is equal to ,
\begin{align*}
& \frac{\partial \log S(\mathbf{x}^{\alpha}, \mathbf{y}^{\alpha})}{\partial s(\mathbf{x}^{\alpha}, \mathbf{y}^{\alpha}, i, j)} \\
&= \frac{1}{S(\mathbf{x}^{\alpha}, \mathbf{y}^{\alpha})} \frac{\partial S(\mathbf{x}^{\alpha}, \mathbf{y}^{\alpha})}{\partial s(\mathbf{x}^{\alpha}, \mathbf{y}^{\alpha}, i, j)} \\
&=\frac{1}{S(\mathbf{x}^{\alpha}, \mathbf{y}^{\alpha})} \sum_{\mathbf{z}} e^{s(\mathbf{x}, \mathbf{y}, \mathbf{z})} \frac{\partial s(\mathbf{x}, \mathbf{y}, \mathbf{z}) }{\partial s(\mathbf{x}^{\alpha}, \mathbf{y}^{\alpha}, i, j)}\\
&= \sum_{\mathbf{z}} \frac{e^{s(\mathbf{x}, \mathbf{y}, \mathbf{z})}}{S(\mathbf{x}^{\alpha}, \mathbf{y}^{\alpha})}  \mathbbm{1}[(i, j) \in \mathbf{z} ]\\
&=  \sum_{\mathbf{z}} P(\mathbf{z}|\mathbf{x},\mathbf{y})  \mathbbm{1}[(i, j) \in \mathbf{z} ]\\
&=  \mathbbm{E}_{\mathbf{z} \sim P(\mathbf{z}|\mathbf{x},\mathbf{y})} \mathbbm{1}[(i, j) \in \mathbf{z} ]\\
\end{align*}

which can be exactly computed by the inside-outside algorithm \cite{baker1979trainable,eisner-1996-coling,paskin2001cubic}.
Term 2 can be computed using any automatic differentiation toolkit, such as those in TensorFlow and PyTorch. 

In a similar manner, the gradient of the partition function can also be exactly computed using a modified inside-outside algorithm (by running the inside-outside algorithm on the WCFG introduced in section \ref{sec:model}) and automatic differentiation. Therefore, we can perform end-to-end parameter learning with gradient descent.

\subsection{Limiting Dependency Length}
The NLDM can capture dependency relations between labels of arbitrary distance. In most cases, however, very long dependencies are rare and therefore we can set an upper-bound  $k$ of the dependency length. The time complexity of both the inside-outside algorithm and the CYK algorithm can then be reduced to $O(m^2 \times n \times k^2)$.

\section{Relations to Previous Work on Sequence Labeling}

\begin{figure}[tb]
	\centering
	\includegraphics[scale=0.2]{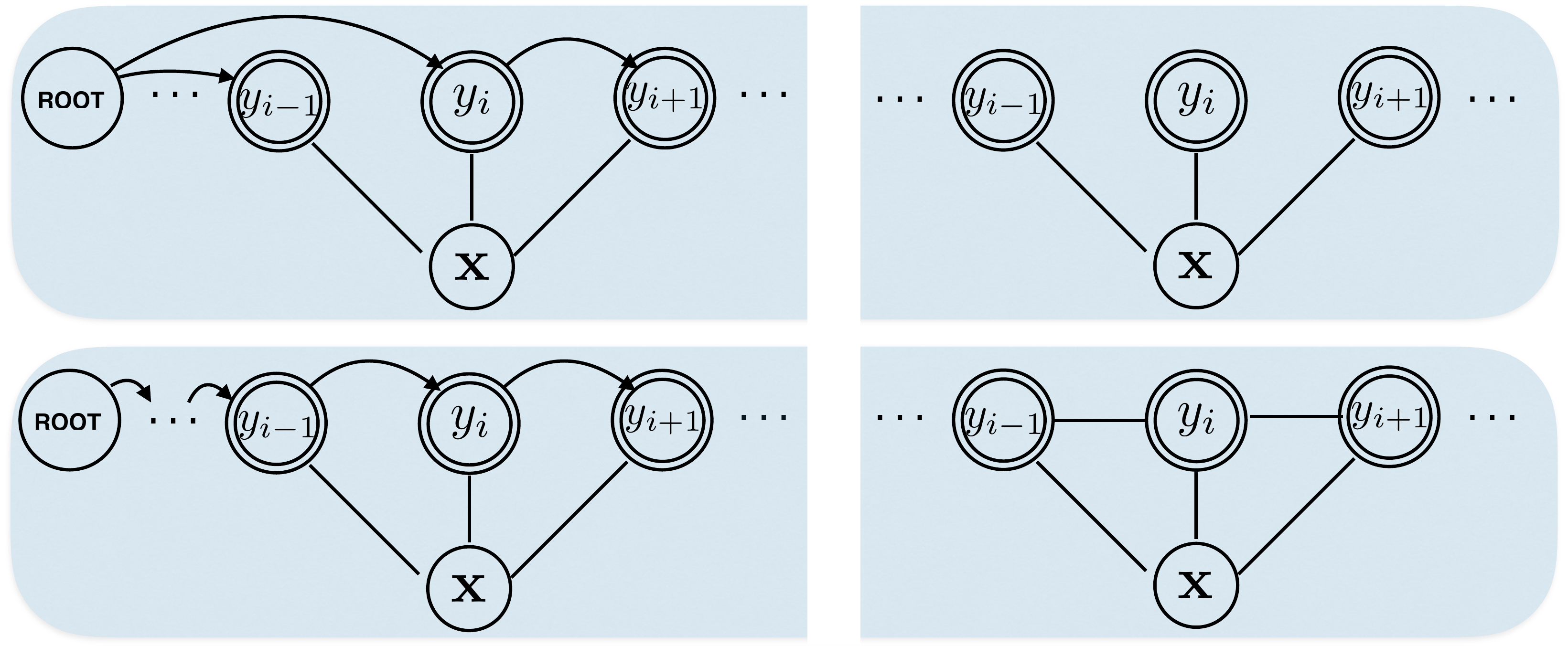}
	\caption{\textbf{Top}: The neural softmax model shown as a special case of the NLDM (left) and the Markov network (right). \textbf{Bottom}: The linear chain CRF shown as a special case of the NLDM (left) and the Markov network (right).}
	\label{fig:comps}
\end{figure}

We show that the neural softmax model and the linear chain neural CRF can be seen as special cases of NLDMs, which is illustrated in Figure \ref{fig:comps}. 

\paragraph{Relation to Neural Softmax} 

In a NLDM, if we only allow dependencies from the root label $y_0$ to the other labels, then there is only one possible tree. Based on the transition-emission edge score formulation, the score of the $(\mathbf{x}, \mathbf{y})$ pair becomes,
\begin{align*}
S(\mathbf{x}, \mathbf{y}) &= \sum_{\mathbf{z}} e^{s(\mathbf{x}, \mathbf{y}, \mathbf{z})}
= e^{\sum_{i=1}^n s(\mathbf{x}, \mathbf{y}, 0, i)}
\\ &= e^{\sum_{i=1}^n \phi (\mathbf{x}, y_i, i) + \psi(y_{0}, y_{i})}
\end{align*}
If we further assume that $\psi(y_{0}, y_{i})$ is always 0, then we have
\begin{align*}
S(\mathbf{x}, \mathbf{y}) = e^{\sum_{i=1}^n \phi (\mathbf{x}, y_i, i)} = e^{\sum_{i=1}^n \mathbf{h}_{i}^{T} \mathbf{W}_{e} \mathbf{t}_{y_{i}}}
\end{align*}

Then the conditional probability of the labels given the input sequence becomes,
\begin{align*}
P(\mathbf{y}|\mathbf{x}) &= \frac{e^{\sum_{i=1}^n \phi(\mathbf{x}, y_i, i)}}{\sum_{\mathbf{y'}}e^{\sum_{i=1}^n \phi(\mathbf{x}, y'_i, i)}} \\ &= \frac{\prod_{i=1}^{n} e^{\phi(\mathbf{x}, y_i, i)}}{\prod_{i=1}^{n} \sum_{y'} e^{ \phi(\mathbf{x}, y', i)}} \\
&= \prod_{i=1}^{n} \frac{e^{\phi(\mathbf{x}, y_i, i)}}{\sum_{y'} e^{ \phi(\mathbf{x}, y', i)}} 
\end{align*}
We can see that in this case the NLDM becomes equivalent to a neural softmax model described in Section \ref{sec:2.2}.

\paragraph{Relation to Linear Chain Neural CRFs} 

In a NLDM, if we only allow dependencies from every label to its next label, then again there exists only one possible tree. Based on the transition-emission edge score formulation, the score of the $(\mathbf{x}, \mathbf{y})$ pair becomes,
\begin{align*}
    &S(\mathbf{x}, \mathbf{y})= \sum_{\mathbf{z}} e^{s(\mathbf{x}, \mathbf{y}, \mathbf{z})}
    = e^{\sum_{i=1}^n s(\mathbf{x}, \mathbf{y}, i-1, i)} \\
    &= e^{\sum_{i=1}^n \left( \phi (\mathbf{x}, y_i, i) + \psi(y_{i-1}, y_{i}) \right)}
\end{align*}
which is exactly the linear chain neural CRF model illustrated in Section \ref{sec:2.1}. 

\section{Experiments}
We evaluate our model on both synthetic and real data.
For comparison, we choose the following baseline models for sequence labeling. 

\paragraph{Neural Softmax} Neural Softmax is a simple and strong baseline in sequence labeling. It considers labeling sequence as several independent supervised classification problems. In our experiments, we used the same feature extraction layer for comparison.

\paragraph{Neural CRFs} Neural CRFs are important baselines since our model can be seen as their extension. We use 1/2-order CRFs as our two CRFs baselines. In our experiments, we used the same feature extraction layer for neural CRFs and NLDMs in order to make a fair comparison. We also perform the trilinear function for neural CRFs models.

We use a Bi-LSTM as the neural encoder in all the models. The code of our approach and the baseline systems, the hyper-parameters, and the trained models will be released at \url{https://anonymous.link}.

\subsection{Hyper-parameters}
For all the models, we tune the learning rate in the range of \{0.03, 0.1, 0.3\} and the LSTM hidden size $D_h$ in the range of \{100, 200, 400\}. For CRF and NLDM models, we select the label embedding size $D_t$ from \{10, 50, 100\}, tensor rank $D_r$ from \{100, 400, 600\}, and the dependency length limit $k$ from \{2, 3, \dots, 10\}.

\subsection{Synthetic data}
We compared the NLDM and the neural CRF on a synthetic dataset generated from a novel infinite order hidden Markov model in which, at each step, the transition distribution is computed by an LSTM that takes the existing label sequence as input, and the emission distribution is computed by another LSTM that takes the existing word sequence as input.
To generate the dataset, we set the label number to 5, word dictionary size to 1000, and maximum sequence length to 10. For the LSTMs, the size of both the hidden state and the output state is set to 50 while the parameters are randomly generated. The embeddings of the labels and words are also randomly generated.  We generated 1k, 3k, 10k, 30k, and 100k samples and randomly selected 80\% for training, 10\% for validation and 10\% for testing. 
For all the experiments, we did not tune the hyper-parameters and kept them the same in the neural CRF and NLDM.

\begin{figure}[t]
\centering
\includegraphics[scale=0.5]{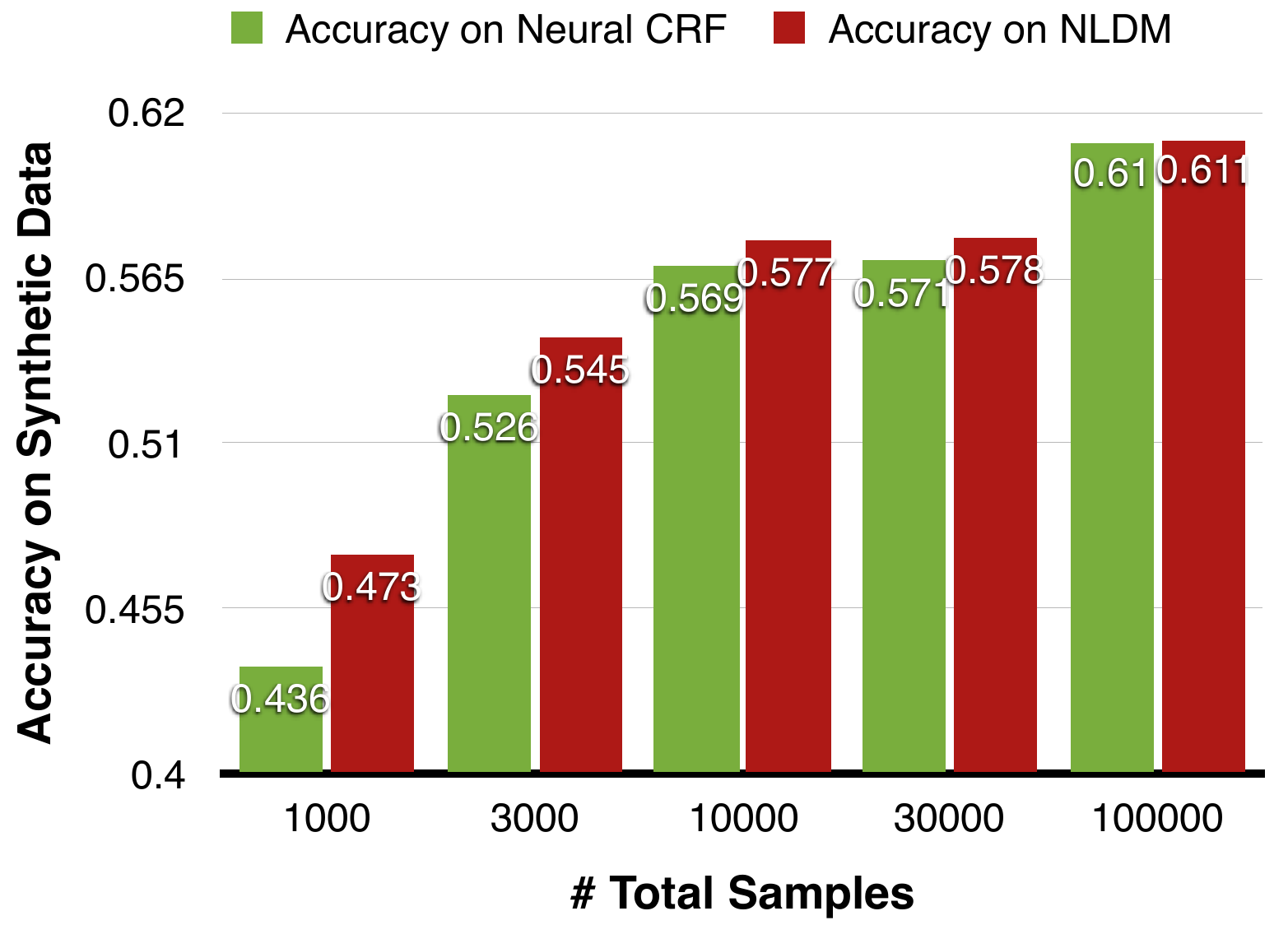}
\caption{Results on the synthetic data. The neural CRF and the NLDM utilize the Bi-LSTMs as the neural encoder.}
\label{fig:syn-results}
\end{figure}

The experimental results are shown in Figure \ref{fig:syn-results}. As can be seen, with more training samples, the performances of both the neural CRF and the NLDM increase. When the sample size is small, the NLDM has significant advantage over the neural CRF model. For larger sample sizes, the NLDM still achieves better accuracies although with smaller margins.
\begin{table*}
\setlength\tabcolsep{4.3pt}
\renewcommand\arraystretch{1.16}
	\centering
	\begin{tabular}{cccccccccc}
		\hline
		{\bf } & {\bf\textsc{Tw}} &{\bf\textsc{En}}& {\bf\textsc{Fr}}& {\bf\textsc{It}}& {\bf\textsc{Te}}& {\bf\textsc{Pl}}& {\bf\textsc{Mt}}& {\bf\textsc{Ru}}& {\bf\textsc{Sv}} \\\hline
        \# Vocab  &4806 & 94217& 35373 & 124445&6465&130967&44162&98000&90960\\
		\# Labels  & 25 &  17&  15 & 16&14&15&17&16&17\\
		\# Training & 1327 & 3176 &  1153 & 5368&1051&13774&1123&3850&2738\\
        \# Dev & 500 & 1032& 907& 671&131&1745&433&579&912\\
        \# Test & 547 & 1035&726& 674&146&1727&518&601&914\\ \hline
	\end{tabular}
	\caption{Statistics of the nine sequence labeling datasets.}
    \label{table:pos-data}
\end{table*}

\begin{table*}
    \centering
    \setlength\tabcolsep{5.3pt}
\renewcommand\arraystretch{1.16}
    \begin{tabular}{ccccccccccc}
    \hline
              & {\bf\textsc{Tw}} & {\bf\textsc{En}}& {\bf\textsc{Fr}}& {\bf\textsc{It}}& {\bf\textsc{Te}}& {\bf\textsc{Pl}}& {\bf\textsc{Mt}}& {\bf\textsc{Ru}}& {\bf\textsc{Sv}}& {\bf\textsc{Avg.}} \\ \hline 
    Neural Softmax &  88.73&90.22 & 89.24 & 89.70 & 82.79 & 89.02 & \bf89.16 & 82.78 & 89.50&87.90\\ 
    Neural 1-order CRF & 89.09 & 90.22 & 89.24 & 89.71 & 83.36 & 89.01 & 88.89 & 82.79 & 89.39&87.97\\
    Neural 2-order CRF &   89.29   &90.18 & 89.45 & 89.71 & 85.35 & \bf89.22 & 89.13 & 82.95 & 89.48   &   88.31   \\ 
    NLDM & {\bf 89.37} &{\bf90.41} & \bf89.64 & \bf90.58 & \bf86.45 & 89.18 & 89.11 & \bf83.23 & \bf89.77   &  \bf88.64    \\ \hline
    \end{tabular}
\caption{POS tagging accuracy.}
\label{table:twitter-result}
\end{table*}

\subsection{Real data}\label{sec:real-data}
We performed experiments on the POS tagging task. We used the Twitter datasets from \newcite{gimpel2011part} and \newcite{owoputi2013improved}. We merged the 1000-tweet OCT27-TRAIN set and the 327-tweet OCT27-DEV set as our training dataset, used the 500-tweet OCT27-TEST set for validation, and used the 547-tweet DAILY547 test set for testing. We also selected eight languages from Universal Dependencies tree banks (UD). The statistics of the datasets are showed in Table \ref{table:pos-data}. 
We report the experimental results in Table \ref{table:twitter-result}. It can be seen that our model achieves the best results in most of the experiments.

\begin{table}
    \centering
\begin{tabular}{ccc}
    \hline {\bf\textsc{Dependency Len}} & {\bf\textsc{Freq.}} & {\bf\textsc{Freq. (gold)}}\\ \hline 
    1 & 	65.3 & 65.4\\ 
    2-10 & 	20.7 & 34.3\\ 
    $>$ 10 & 	14.0 & 0.3\\ \hline
\end{tabular}
\captionof{table}{Frequency (\%) of different dependency lengths in parses decoded by our model and in gold parses on the Twitter dataset.}
\label{table:freq-edge-tree}
\end{table}

\section{Analysis}
\subsection{Impact of Dependency Length Limit}
\begin{figure}
\centering
\includegraphics[scale=0.45]{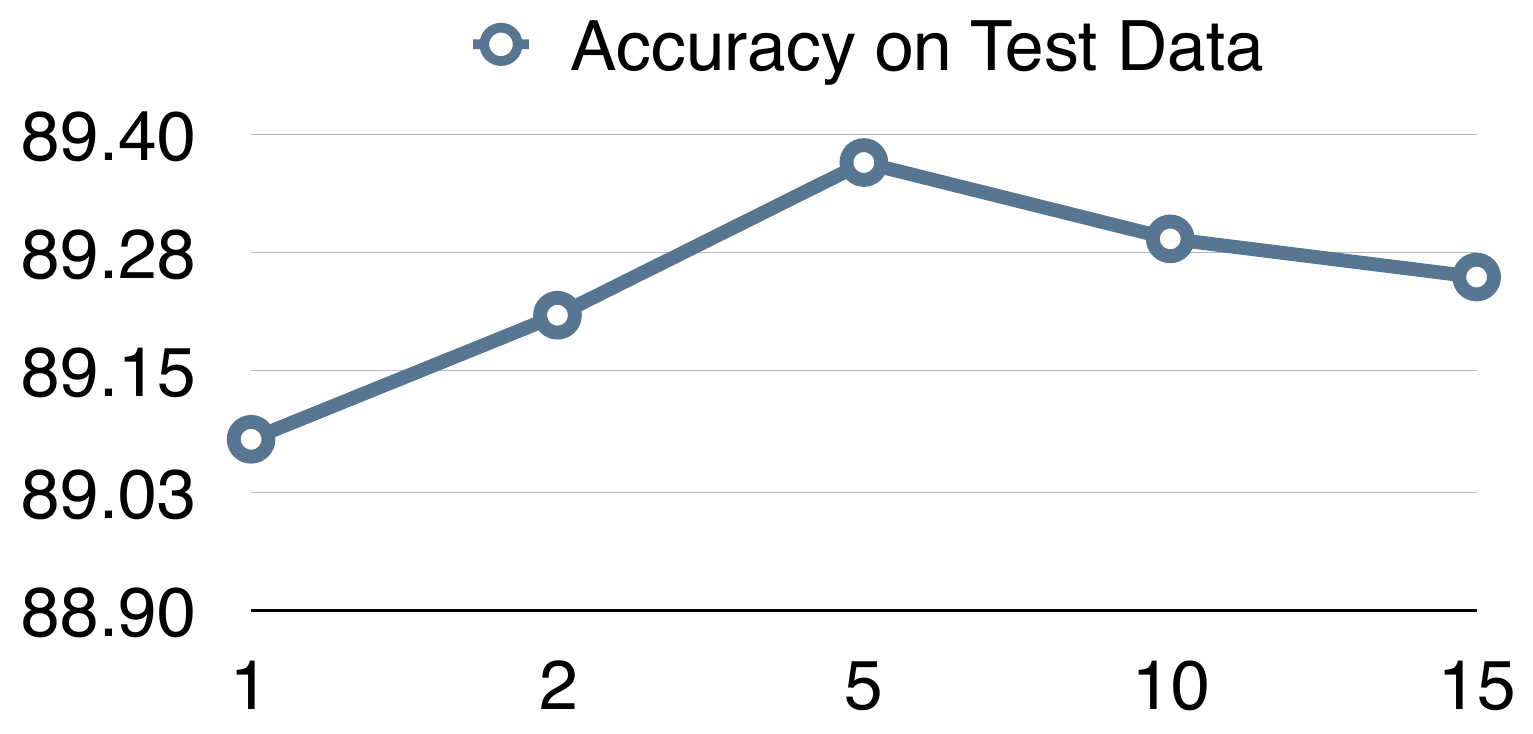}
\captionof{figure}{Accuracy on the Twitter test set with different dependency length limits.}
\label{fig:tree-acc}
\end{figure}

We analyze the performance of our model on the Twitter POS tagging dataset with different values of the dependency length limit $k$. We repeat the experiments in Section \ref{sec:real-data} with $k\in\{1,2,5,10,15\}$ and report the results in Figure \ref{fig:tree-acc}.
Note that $k=1$ corresponds to linear chain neural 1-order CRFs. It can be seen that allowing long-range dependencies is indeed beneficial compare with the linear chain baseline. However, setting the $k$ value beyond 5 seems to decrease the accuracy, possibly because dependencies longer than 5 are relatively rare on the dataset.

\subsection{Analysis of Induced Tree Structures}
We use the learned NLDM from Section \ref{sec:real-data} as a dependency parser by solving $\mathbf{z}^*=\arg\max_{\mathbf{z}} \max_{\mathbf{y}} s(\mathbf{x}, \mathbf{y}, \mathbf{z})$ and compare the dependency trees decoded by the NLDM with the gold linguistic trees on the Twitter dataset.

\paragraph{Comparison on Dependency Length} 
We first aim to analyze whether the NLDM indeed learns to capture long-range dependency relations. We count the frequencies of different dependency lengths in the decoded dependency parses and report the results in Table \ref{table:freq-edge-tree}. From the table we can see that over 30 percent of the dependencies are longer than 1, suggesting that our model is able to capture long-range dependencies. In addition, the NLDM can be seen to induce longer dependencies compared with the gold parses.

\paragraph{Comparison with Gold Trees} 
We compare the dependency trees decoded from our model with the gold linguistic dependency trees from the Tweebank \cite{kong-EtAl:2014:EMNLP2014}. The unlabeled dependency accuracy of our model is 17.9\%, which suggests that our model induces dependency trees that are significantly different from the gold trees, echoing the findings from previous work \cite{yogatama2016learning,williams2017learning}.

\section{Conclusion}
We propose a novel latent variable probabilistic model for sequence labeling called the Neural Latent Dependency Model (NLDM). A NLDM allows dependencies between labels of arbitrary length and models the set of dependencies with a latent projective tree structure. We prove that the neural softmax model and the linear chain neural CRF model are two special cases of our model. We present an end-to-end training algorithm and an efficient approximate decoding algorithm. We tested our model on both synthetic and real datasets and show that our model achieves superior performance over baselines. For future work, we plan to design better decoding algorithms and apply our model to other more complicated sequence labeling tasks such as semantic role labeling.

\section*{Acknowledgement}
This work was supported by Alibaba Group through Alibaba Innovative Research Program. This work was also supported by the National Natural Science Foundation of China (61976139).

\bibliography{nldm}
\bibliographystyle{acl_natbib}
\end{document}